\documentclass[12pt,letterpaper]{article}

\usepackage[left = 1 in, right = 1 in, top = 1 in, bottom = 1 in]{geometry}
\usepackage{algorithm}% http://ctan.org/pkg/algorithm
\usepackage[noend]{algpseudocode}% http://ctan.org/pkg/algorithmicx
\usepackage{array}
\usepackage{adjustbox}
\usepackage{makecell}
\usepackage{amsmath}
\usepackage{amssymb} % symbols and math
\usepackage{graphicx}
\usepackage{setspace}
\usepackage{float}        % For figures
\usepackage{siunitx}
\usepackage{xspace}
\usepackage{xcolor}
\usepackage{listings} % Used for code
\usepackage{upgreek}
\usepackage{array}
\newcolumntype{L}{>{\raggedright\arraybackslash}m{4.7cm}}

\newcommand{\reffig}[1]{Fig.~\ref{#1}}
\newcommand{\refsec}[1]{Section~\ref{#1}}

\newcommand{\refeq}[1]{Eq.~\ref{#1}}
\newcommand{\refcite}[1]{Ref.~\cite{#1}}
\newcommand{\reftab}[1]{Table~\ref{#1}}
\newcommand{\refalg}[1]{Algorithm~\ref{#1}}

\newcommand{\ttheta}[0]{\ensuremath{\tilde{\theta}}\xspace}
\newcommand{\tc}[0]{\ensuremath{\tilde{C}}\xspace}
\newcommand{\yhat}[0]{\ensuremath{\hat{y}}\xspace}
\newcommand{\tautheta}{\ensuremath{\tau_\uptheta}\xspace}
\newcommand{\tauhp}{\ensuremath{\tau_{\text{hp}}}\xspace}
\newcommand{\taup}{\ensuremath{\tau_\text{p}}\xspace}
\newcommand{\taux}{\ensuremath{\tau_\text{x}}\xspace}
\newcommand{\Cnoise}{\ensuremath{C_{\text{noise}}}\xspace}
\newcommand{\thetanoise}{\ensuremath{\theta_{\text{noise}}}\xspace}
\newcommand{\sigmac}{\ensuremath{\sigma_{\text{C}}}\xspace}
\newcommand{\sigmatheta}{\ensuremath{\sigma_{\uptheta}}\xspace}

\graphicspath{ {./fig} }

\begin{document}

\doublespacing
\pagenumbering{arabic} % Adds page numbers like "1,2,3"

%%%%%%%%%%%%%%%%%%%%%%%%%%%%%%%%%%%%%%%%%%%%%%%%%%%
%%% Basic title - Draft name, Authors, Affiliations %%%
%%%%%%%%%%%%%%%%%%%%%%%%%%%%%%%%%%%%%%%%%%%%%%%%%%%

\title{Multiplexed gradient descent: Fast online training of modern datasets on hardware neural networks without backpropagation}
\author{A. N. McCaughan$^1$, B. G. Oripov$^2$, N. Ganesh$^1$, S. W. Nam$^1$,\\ A. Dienstfrey$^1$,  S. M. Buckley$^1$}

\date{% Cheap hack to add affiliations without any special packages
    \small
    $^1$National Institute of Standards and Technology, Boulder, CO 80305\\%
    $^2$University Colorado Boulder, Boulder, CO 80309\\%
}
\maketitle

\begin{abstract}
    We present multiplexed gradient descent (MGD), a gradient descent framework designed to easily train analog or digital neural networks in hardware. MGD utilizes zero-order optimization techniques for online training of hardware neural networks. We demonstrate its ability to train neural networks on modern machine learning datasets, including CIFAR-10 and Fashion-MNIST, and compare its performance to backpropagation. Assuming realistic timescales and hardware parameters, our results indicate that these optimization techniques can train a network on emerging hardware platforms orders of magnitude faster than the wall-clock time of training via backpropagation on a standard GPU, even in the presence of imperfect weight updates or device-to-device variations in the hardware. We additionally describe how it can be applied to existing hardware as part of chip-in-the-loop training, or integrated directly at the hardware level.  Crucially, the MGD framework is highly flexible, and its gradient descent process can be optimized to compensate for specific hardware limitations such as slow parameter-update speeds or limited input bandwidth.
\end{abstract}

%%%%%%%%%%%%%%%%%%%%%
%%% Document text %%%
%%%%%%%%%%%%%%%%%%%%%

\section{Introduction}
\label{sec:introduction}

Machine learning has proven an invaluable tool for a variety of applications \cite{LeCun2015}. However, machine learning on traditional digital hardware is inefficient, leading to a significant effort towards building custom hardware that can perform machine learning tasks at high speeds with lower energy costs \cite{Schuman2017}.  A number of hardware platforms have emerged using analog \cite{Haensch2019}, digital \cite{Merolla2014, Davies2018}, or mixed-signal processing \cite{Meier2015} that will potentially offer increased operational speeds and/or reduced energy costs \cite{Mehonic2022}. However, many of the most promising hardware instantiations only perform the inference part of the machine learning algorithm. Meanwhile the larger portion of the energy cost is spent training on datasets \cite{Thompson2020}, usually via gradient descent. Backpropagation is by far the most commonly used method of computing the gradient for gradient descent, but has proved to be challenging to implement in novel hardware platforms \cite{Wright2022}.

Though often conflated, training via gradient descent does not require backpropagation – backpropagation is only used to calculate the gradient. Other methods for computing the gradient in neural networks exist, but are much less efficient in software than backpropagation and so are rarely used in today's machine learning applications. This is not generally true in hardware, where backpropagation may not only be challenging to implement, but also may not be the most efficient way to compute the gradient.

Of particular interest in hardware are model-free methods, in which we require no knowledge of the internal structure of the network (e.g topology, activation function, derivatives, etc), only the ability to perturb the network's parameters and measure the network's response. The simplest example of such a method is finite-difference \cite{Kiefer1952}, which has been employed for chip-in-the-loop training \cite{Shen2017}. However, finite-difference has several other disadvantages that prevent its widespread implementation in hardware, including the requirements for extra memory at every synapse and global synchronization. Fortunately, there are a variety of other model-free methods that overcome some of the issues associated with finite-difference \cite{Spall1992, Dembo1990}.

In this paper, we show that model-free perturbative methods can be used to efficiently train modern neural network architectures in a way that can be implemented natively within emerging hardware. These methods were investigated for training VLSI neural networks beginning in the 1990s \cite{Matsumoto1990, Jabri1992, Alspector1992, Kirk1992, Maeda1995, Cauwenberghs1996, Miyao1997,   Montalvo1997, Maeda2003, Adhikari2015}, and more recently on memristive crossbars \cite{Wang2018} and photonic hardware \cite{Bandyopadhyay2022}, but all these demonstrations have been very limited in scale, comprising small datasets with only a few neurons.  Below we describe a framework for applying these techniques to existing hardware at much larger scales, with an emphasis on creating simple, highly-localized circuits that could be implemented on-chip if desired. The framework is also extensible to training existing hardware systems via a chip-in-the-loop technique. We note that these methods have also been adapted in forward gradient approaches using auto-differentiation, which have attracted recent interest in the machine learning literature \cite{Baydin2022,Silver2022,Ren2023}.

We show that under realistic assumptions of the operating timescales of analog and digital hardware neural networks, one can train hardware to solve modern datasets such as CIFAR-10 faster than training a software network on a GPU, even in the presence of signal noise and device-to-device variations in the hardware. A major advantage of this framework is that it can be used to perform online training of hardware platforms originally designed only for inference while making minimal hardware modifications.

\section{Multiplexed gradient descent}
\label{sec:multiplexed}

\subsection{Computing the gradient with perturbations}
\label{sec:computing}

\begin{figure}[h] % [H] forces the figure to appear exactly here
    \centering
    \includegraphics[width=6.5in]{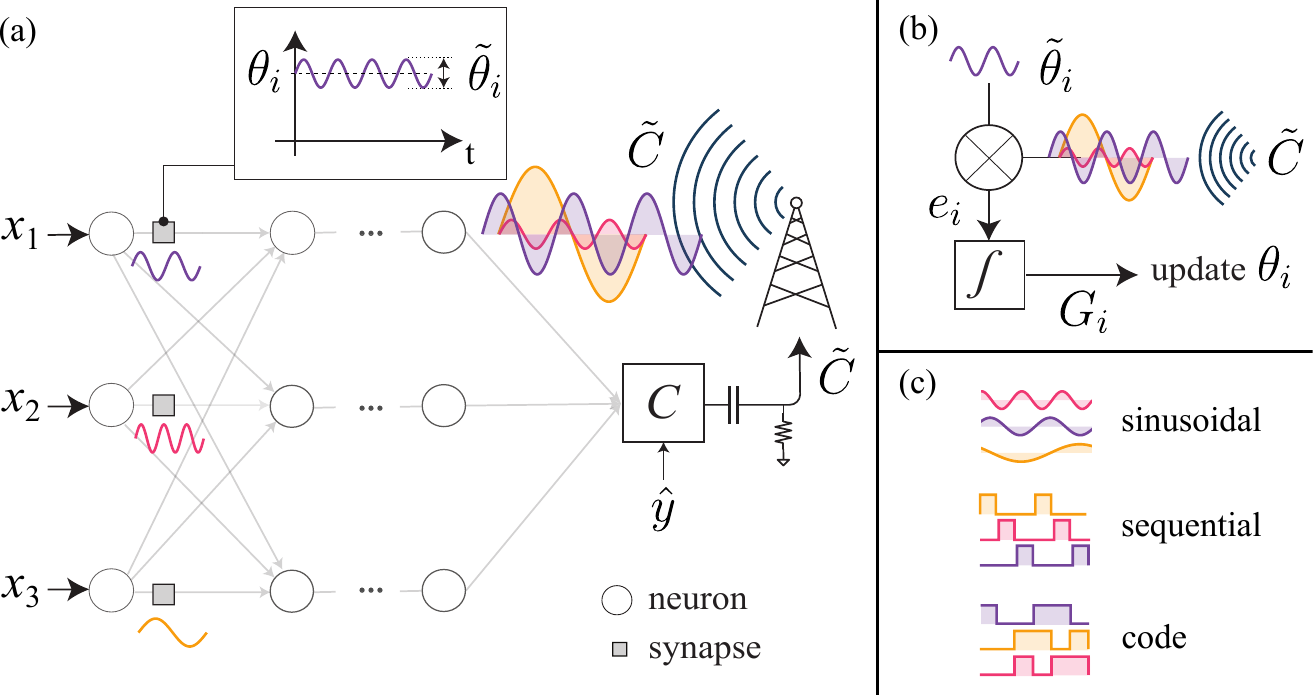}
    \caption{ (a) Schematic diagram showing the operation of the MGD framework in a feedforward neural network using example sinusoidal perturbations.  (a, inset) Each parameter $i$ is modulated slightly from its base value $\theta_i$ by the perturbation $\ttheta_i$. The result of these perturbations causes a modulation in the cost $\tc$, which is globally broadcast back to all the parameters. (b) A homodyne detection process is used to compute the partial gradient approximations $G_i$ from the integrated product of $\theta_i$ and $\tc$. This partial gradient is then used to update $\theta_i$ in the approximate direction of the gradient. (c) Example perturbation types that can be used with this process.}
       \label{fig:1-mgd}
\end{figure}

We begin with the basic assumption that we have some hardware with programmable parameters (e.g. weights and biases) that can perform inference. Our goal is to augment the hardware minimally such that it can also be trained via gradient descent. We will show how to configure the hardware such that the network as a whole automatically performs gradient descent, without backpropagation. As an example, assume we have a hardware instantiation of a feedforward multi-layer neural network as shown in  \reffig{fig:1-mgd}. The hardware takes time-varying inputs $x(t)$, training target $\yhat(t)$, has variable parameters $\theta$, outputs the inference $y(t)$, and computes a cost $C(y(t), \yhat(t))$. To allow us to compute the gradient of such a system, we first add a small time-varying perturbation $\ttheta_i(t)$ to each parameter base value $\theta_i$ (\reffig{fig:1-mgd}a, inset).  This perturbation will slightly modulate the cost $C$, and that modulation will be fed back to the parameters.  This process will ultimately allow us to extract the gradient of the system.

Although the perturbations can take a variety of different forms, we will first describe this process by using sinusoidal perturbations as they are conceptually straightforward to understand. In this scenario, each parameter $\theta_i$ is slightly modulated at a unique frequency $\omega_i$ and amplitude $\Delta \theta$, giving the perturbation $\tilde{\theta}_i(t) = \Delta \theta \sin(\omega_i t)$.  As each parameter is modulated, it slightly changes $y(t)$ which in turn changes the cost. Thus, if the parameters are modulated by frequencies $\omega_1$, $\omega_2$, $\omega_3$, etc, those same frequencies will necessarily appear as small modulations in the cost $\tc(t)$ on top of the baseline (unperturbed) cost value $C_0$, such that

\begin{equation}
\label{eqn:C}
C(t) = C_0 + \tc(t) = C_0 + \sum_i \Delta C_i \sin(\omega_i t)
\end{equation}

If we remove $C_0$, we are left with a time varying signal $\tc(t) = \sum_i \Delta C_i \sin(\omega_i t)$ corresponding only to the effects of our parameter perturbations. The amplitude $\Delta C_i$ is simply the amplitude of change in the cost due to $\ttheta_i(t)$, the perturbation of parameter~$i$.

Since the gradient with respect to the cost $dC/d\theta$ is composed solely from the partial gradients $dC/d \theta = (\partial C/\partial \theta_1, \, \partial C/\partial \theta_2, \, ...)$, if we can extract $\Delta C_i$ for each parameter we can produce an estimate of the complete gradient $G =(\Delta C_1 / \Delta \theta_1, \Delta C_2 / \Delta \theta_2, ...)$. Now the task becomes to extract individual $\Delta C_i$ out of the summed signal $\tc(t)$. Fortunately, to extract a given $\Delta C_i$, all we need to do is integrate the product of the input perturbation $\ttheta_i(t)$ with $\tc(t)$. The integration takes the form of a homodyne detection, where unwanted perturbations (frequencies) from other parameters are eliminated via integration:

\begin{equation}
\label{eqn:Gsine}
\begin{split}
G_i &= \frac{1}{\Delta \theta_i^2}\frac{1}{T}\int_{t=0}^T \sum_{k} \Delta C_k \sin(\omega_k t) \Delta\theta_i \sin(\omega_i t) dt \\ 
&= \frac{\Delta C_i}{\Delta \theta_i} \quad \textrm{as} \quad T \rightarrow \infty \\
\end{split}
\end{equation}
 where $1/\Delta \theta_i^2$ is a normalization constant.

The value $G_i$ is the approximation for the partial gradient for parameter $i$.  $G$ approaches the exact gradient when both $T \rightarrow \infty$ and the amplitude of the perturbation $\Delta \theta_i$ approaches zero, and is only an approximation otherwise.  Fortunately, even at realistic timescales and amplitudes, $G$ contains meaningful information and can be used to perform gradient descent~\cite{Spall1992}.

For illustrative purposes we have described the algorithm using sinusoidal parameter perturbations. However, any collection of orthogonal, mean zero perturbations can be used \cite{Dembo1990}, including a variety of analog and discrete perturbations as shown in \reffig{fig:1-mgd}c.  In general, we will be integrating the product $e_i(t) = \tc (t) \ttheta_i (t) /\Delta \theta_i^2$, which we refer to as the error signal, and $G_i$ will be given by \footnote{Note that here and in the simulation results, $G_i$ is being accumulated with time and is not normalized by $1/T$, unlike \refeq{eqn:Gsine}. As described later, this allows us to vary the integration time without greatly impacting the rate of training--equivalently, one can  integrate for a long time resulting in a large step down the gradient, or one can take a series of shorter steps instead and travel approximately the same distance along the gradient.}

\begin{equation}
\label{eqn:G}
G_i = \int_{t=0}^T \frac{\tc(t) \ttheta_i(t)}{\Delta \theta_i^2} dt 
\end{equation}

 We discuss the effects of changing the perturbation type in \refsec{sec:simulation:analog}.  We also note that although many of the quantities described here are time-varying, in the following sections we will drop the explicit time dependence notation for the sake of brevity.

\subsection{Gradient descent in the MGD framework}
\label{sec:gradient}

Here we describe the practical implementation of a model-free gradient descent framework in hardware, which we term multiplexed gradient descent (MGD). To better understand the algorithm from a hardware perspective, we will run through the same computation previously described, but from the viewpoint of a single parameter (e.g. a synapse weight in a hardware neural network). The process begins with the application of a local perturbation $\ttheta_i$ that slightly modifies the base value of the parameter $\theta_i$ (\reffig{fig:1-mgd}a, inset). As previously described, this perturbation -- and any other perturbations from other parameters -- induce a change in the cost $\tc$ on top of the baseline cost $C_0$ such that the cost at the output is $C = C_0 + \tc$. $\tc$ may be extracted from $C$ either by direct subtraction of $C_0$ or, in some analog scenarios, by a simple highpass filter. The resulting $\tc$ signal is broadcast globally to all parameters, so our parameter $\theta_i$ has access to it. (Note that although \reffig{fig:1-mgd} shows a wireless broadcast tower for purposes of clarity, in most hardware platforms this will be a wired connection). However, we must assume that parameters other than the $i$th are also causing modulations in the cost as well.  To our parameter $\theta_i$, these other modulations are unwanted and must be filtered out. As shown in \reffig{fig:1-mgd}b, for the parameter $i$ to extract only its own effect on the cost, it can just integrate the product of its local perturbation $\ttheta_i$ and the global cost signal $\tc$ it receives. This has the effect of isolating the contribution from $\theta_i$ due to the pairwise orthogonality of the perturbation signals. From \refeq{eqn:G}, this integration produces the partial gradient approximation $G_i \propto \Delta C_i / \Delta \theta_i$. The parameter can then use the $G_i$ value to directly to reduce the cost by updating itself according to a gradient descent step

\begin{equation}
\label{eqn:update}
\theta_i \rightarrow \theta _i- \eta G_i
\end{equation}

where $\eta$ is the learning rate. When all the parameters of a system perform this operation in parallel, the resulting weight update corresponds to gradient descent training of the entire network. Because all the parameters are being perturbed and updated simultaneously, we call the framework multiplexed.

Looking at this process from the hardware perspective, one must also examine several practical considerations such as when to perform the weight update, how long to integrate the gradient, and so forth. We introduce the following time constants to provide a framework for managing these considerations in the MGD context.

% These considerations are addressed in the MGD framework using the time constants \tautheta, \taup, and \taux. 
\begin{itemize}
    \item  \taup is the timescale over which perturbations occur.  In a digital system, the perturbations of each parameter would be updated to new values every \taup.  In an analog (continuous) system, \taup corresponds to the characteristic timescale of the perturbations.  For instance, if using sinusoidal perturbations, it corresponds approximately to the total bandwidth the frequencies occupy.
    \item \tautheta is the gradient-integration time (i.e. $T$ in \refeq{eqn:G}).  It sets how often parameter-updates occur and also determines how accurate the gradient approximation will be.  For each time period \tautheta, the gradient approximation $G$ is integrated, and at the end of that period the parameters are updated according to \refeq{eqn:update}.
    \item \taux controls how often new training samples $x$, $\yhat$ are shown to the hardware.  After each \taux period, the old sample is discarded and a new one is applied, generating new outputs $y$ and cost $C$.
\end{itemize}

\begin{figure}[h] % [H] forces the figure to appear exactly here
    \centering
    \includegraphics[width=6.5in]{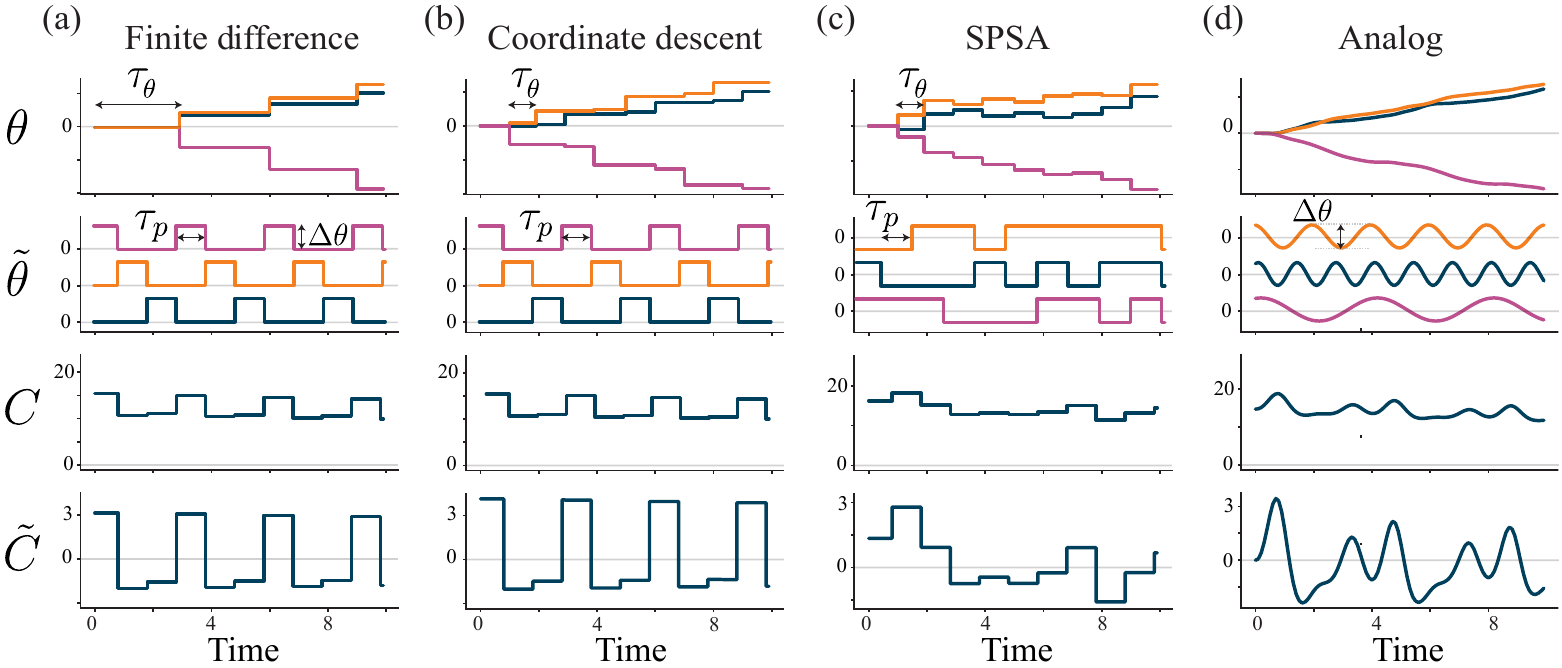}
    \caption{Different optimization algorithms can be implemented in the MGD framework by changing the values the time constants, \taup, \tautheta and \taux, and varying the perturbation type. Shown here are the parameter values $\theta$ that are updated every \tautheta, perturbation \ttheta with characteristic time constant \taup, computed cost $C$, and cost-modulation \tc. Varying \tautheta and \taux and perturbation type allows MGD to implement a variety of optimization algorithms including (a) finite-difference (b) coordinate descent (c) simultaneous perturbation stochastic approximation (SPSA) and (d) an analog implementation. Each line color in the upper two plots corresponds to one of the three parameters in this simulation.}
     \label{fig:2-algtypes}
\end{figure}

The values of the  time constants \tautheta and \taup have a large impact on the particulars of the training, and particular choices of \tautheta and \taup allow the implementation of certain conventional algorithms in numerical analysis. For instance, consider the implementation of the forward finite-difference algorithm within this framework. To do this we start with discrete perturbations, such that every \taup only a single parameter is perturbed by $\Delta\theta$, and the parameters are perturbed sequentially as shown in \reffig{fig:1-mgd}c. When parameter $i$ is perturbed by $\Delta \theta_i$, the cost changes by $\Delta C$ and the resulting partial gradient $  \Delta C / \Delta \theta_i \approx \partial C/\partial \theta_i$ is stored in $G_i$. Now if we set $\tautheta = P \taup$, where $P$ is the number of parameters in the network, this is exactly equivalent to computing a finite-difference approximation of the gradient: each \taup, one element of the gradient is approximated, and after $P \taup$, every partial gradient $G_i$ has been measured and stored.  Since $\tautheta = P \taup$, the weight update only comes after all the partials of $G$ has been collected, just as in finite-difference.
This process is shown schematically in \reffig{fig:2-algtypes}a.

Similarly, other optimization algorithms can be implemented simply by modifying the values of the time constants. For example, using same procedure as above but with the integration time reduced to a single timestep, i.e. $\tautheta = \taup$. This corresponds exactly to coordinate descent. In this case, rather than storing each $G_i$ until all the partials of the gradient are fully assembled, the weight update is applied immediately after each $G_i$ is computed. This may be more appealing from a hardware perspective than a finite-difference approach, as $G_i$ can be used for the weight update and immediately discarded – unlike finite-difference, it does not require a per-parameter memory to store $G_i$ until the weight update is executed. This coordinate-descent process is shown schematically in \reffig{fig:2-algtypes}b.

As a third example, it is possible to implement simultaneous-perturbation stochastic approximation (SPSA) \cite{Spall1992} by only changing the values of the time constants and the form of the perturbation. In this case,  $\tautheta = \taup$ and a random, discrete $\{+\Delta\theta, -\Delta\theta\}$ perturbation is applied to every parameter every \taup, as shown in \reffig{fig:2-algtypes}c. Similar to coordinate descent, this configuration avoids the need for additional per-parameter memories, as $G_i$ values do not need to be stored.  This method avoids the need for global synchronization of the parameters – the perturbations do not need to be sequential, and instead can be generated locally and randomly at the parameter.

In addition to being able to choose these specific optimization algorithms, by varying the time constants $\taup$ and $\tautheta$ one can also implement entirely new optimization algorithms. For instance, in \reffig{fig:2-algtypes}d we show an MGD implementation on an analog system using sinusoidal perturbations that does not correspond to any of the aforementioned methods. In this case, $\taup$ corresponds to the timescale $1/\Delta f$, where $\Delta f$ is the perturbation bandwidth, the difference between the maximum and minimum perturbation frequency. Additionally, in the analog case there is no discrete update of the parameters and instead \tautheta is an integration time constant.  Unlike the discrete case which accumulates $G$ for \tautheta time then resets it to zero, $\theta$ is continuously updated with the output of an lowpass filter with time constant $\tautheta$ (see \refalg{alg:analog}).

\begin{figure}[h] % [H] forces the figure to appear exactly here
    \centering
    \includegraphics[width=2.5in]{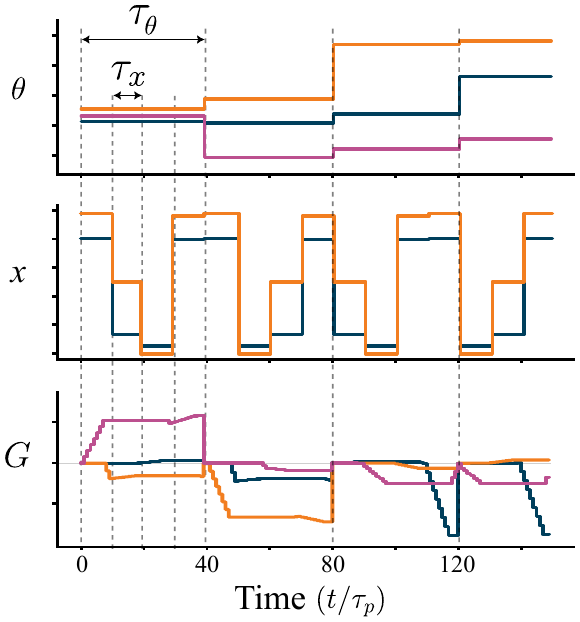}
    \caption{Illustration of batching in a small network with three parameters and two $x$ inputs, training on a dataset with four samples. The parameters $\theta$ are updated every \tautheta, and during that time all four training samples are shown to the network and integrated into the gradient approximation (batch size $\tautheta/\taux = 4$). The gradient approximation $G$ accumulates each timestep, and is reset during the weight-update process after each \tautheta period. Note that the updates to $\theta$ occurs in the opposite direction of $G$, per \refeq{eqn:update}.}
        \label{fig:3-taus}
\end{figure}

Because modern machine learning datasets are composed of many training examples (often tens of thousands), \taux, the time constant that controls how often training samples are changed, is critical. In fact, by setting \taux appropriately, mini-batching can even be implemented in hardware that only allows 1 sample input at a time. The batch size is determined by $\tautheta/\taux$, the ratio of the gradient integration time $\tautheta$ and the sample update time $\taux$. When $\taux$ is shorter than $\tautheta$, multiple samples are shown to the network during a single gradient integration period. As the sample changes, the gradient approximation $G$ will then include gradient information from each of those samples. This integration-in-time process is arithmetically identical to summing multiple examples in parallel (as is often done on a GPU). As a concrete example, for hardware which accepts one input example at a time, if $\tautheta = \taux$ the batch size is 1, but if $\tautheta = 4 \taux$, the batch size is 4. This is shown in \reffig{fig:3-taus} for the same network as in \reffig{fig:2-algtypes} and using simultaneous discrete perturbations. In \refsec{sec:simulation} we demonstrate that batch sizes as large as 1000 function correctly in MGD.

In summary, setting just three time constants and the perturbation type allows the MGD framework to implement a wide range of variations of gradient descent and, depending on the desired approach, one can tailor their training to the needs of the problem and the hardware. For example, MGD can match backpropagation by making \tautheta arbitrarily long\footnote{The MGD approximation to a partial derivative converges to the true partial derivative as \tautheta becomes arbitrarily long. Thus the various parameter update steps presented above can be made arbitrarily close to their analytical definitions as implemented using digital arithmetic and backpropagation.}, as we show in \refsec{sec:simulation:equivalence}. At the same time, we show that short \tautheta values can work equally well. In practice, hardware considerations such as write-speed limitations may require an intermediate \tautheta value, as we discuss in \refsec{sec:hardware}. These features allow the MGD framework to leverage the same training techniques that have been so successful for training machine learning models, while also being compatible with new and emergent hardware.

\section{Simulation of MGD}
\label{sec:simulation}

\subsection{Intro}
\label{sec:simulation:intro}

To characterize the utility of the MGD framework, we first simulated its performance on modern machine learning datasets. The goal of the simulator was not to perform gradient descent as fast as possible on a CPU or GPU, but rather to emulate hardware implementing MGD and evaluate its potential performance in a hardware context. In particular, we used the simulator to estimate the speed, accuracy, and resilience to noise and fabrication imperfections.  The simulator was written in the Julia language and can be run on a CPU or GPU and is available online\footnote{Code available at https://github.com/amccaugh/multiplexed-gradient-descent-paper}. The algorithms used in the simulation are shown in the \refalg{alg:discrete} and \refalg{alg:analog} boxes, with the parameters and their types in software shown in  \reftab{tab:algorithm_parameters}.

\begin{algorithm}
\caption{Discrete algorithm\label{alg:discrete}}
  \begin{algorithmic}[1]
      \State Initialize parameters $\theta$
      \For{$n$ \textbf{in} $\textrm{num\_iterations}$}            
        \If{($n \bmod \taux$ = 0)}
          \State Input new training sample $x$, \yhat
        \EndIf         
        \If{$(n \bmod \taux$ = 0) or ($n \bmod \tautheta$ = 0)}
          \State Set perturbations to zero $\ttheta \leftarrow 0$
          \State Update baseline cost $C_0 \leftarrow C(f(x;\theta), \yhat)$
        \EndIf
        \If{$(n \bmod \taup$ = 0)}
            \State Update perturbations \ttheta
        \EndIf
        \State Compute output $y \leftarrow f(x; \theta+\ttheta)$
        \State Compute cost $C \leftarrow C(y, \yhat)$
        \State Compute change in cost $\tc \leftarrow C-C_0$
        \State Compute instantaneous error signal $e \leftarrow \tc \ttheta /\Delta \theta^2$    
        \State Accumulate gradient approximation $G \leftarrow G + e$            
        \If{($n \bmod \tautheta$ = 0)}
          \State Update parameters $\theta \leftarrow \theta - \eta G$
          \State Reset gradient approximation $G \leftarrow 0$
        \EndIf
      \EndFor
  \end{algorithmic}
\end{algorithm}

\begin{algorithm}
  \caption{Analog algorithm\label{alg:analog}}
  \begin{algorithmic}[1]
      \State Initialize parameters $\theta$
      \For{$t = 0$ \textbf{to} $T$ \textbf{step} $dt$}            
        \If{$(t \bmod \taux$ = 0)}
          \State Input new training sample $x$, \yhat
        \EndIf         
        % \If{$t \bmod \taux$ = 0 or $t \bmod \tautheta$ = 0}
        %   \State Set perturbations to zero $\theta \leftarrow 0$
        %   \State Update baseline cost $C_0 \leftarrow C(f(x;\theta), \yhat)$
        % \EndIf
        \State Update perturbations \ttheta
        \State Compute output $y \leftarrow f(x; \theta+\ttheta)$
        \State Compute cost $C(t) \leftarrow C(y, \yhat)$
        \State Compute discretized highpass $\tc(t) \leftarrow \frac{\tauhp}{\tauhp + dt}\left(\tc(t-dt) + C(t) - C(t-dt) \right)$
        \State Compute instantaneous error signal $e(t) \leftarrow \tc \ttheta dt /\Delta \theta^2$ 
        \State Update gradient approximation $G(t) \leftarrow \frac{dt}{\tautheta + dt} \left( e(t) + \frac{\tautheta}{dt} G(t-dt) \right)$
        \State Update parameters $\theta \leftarrow \theta - \eta G$
      \EndFor
  \end{algorithmic}
\end{algorithm}

\begin{table*}[ht!]
% \begin{adjustbox}{width = 1\textwidth}
\small
    \centering
  \begin{tabular}{L|c|c}
\textbf{Description} & \textbf{Symbol} & \textbf{analog or digital} \\ \hline 
Change in the cost due to perturbation & \tc  & both \\
Perturbation to parameters & \ttheta  & both\\ 
Parameters & $\theta$  & both\\ 
Input sample  & $x$ & both \\ 
Target output & $\yhat$  & both \\
Network output & $y$  & both \\
Cost & $C$  & both \\
Unperturbed baseline cost & $C_0$  & digital \\ 
Gradient approximation & $G$  & both \\ 
Instantaneous error signal & $e$  & both \\ 
Learning rate & $\eta$  & both \\ 
Perturbation amplitude & $\Delta \theta$  & both \\ 
Input-sample change time constant & $\taux$  & both \\
Parameter update time constant & $\tautheta$  & both \\ 
Perturbation time constant & $\taup$ & digital \\ 
Highpass filter time constant & $\tauhp$ & analog \\

\end{tabular}
  % \end{adjustbox}
  \caption{Algorithm parameters and variables used in the simulations}
    \label{tab:algorithm_parameters}
\end{table*}

\begin{figure}[h] % [H] forces the figure to appear exactly here
    \centering
    \includegraphics[width=6.5in]{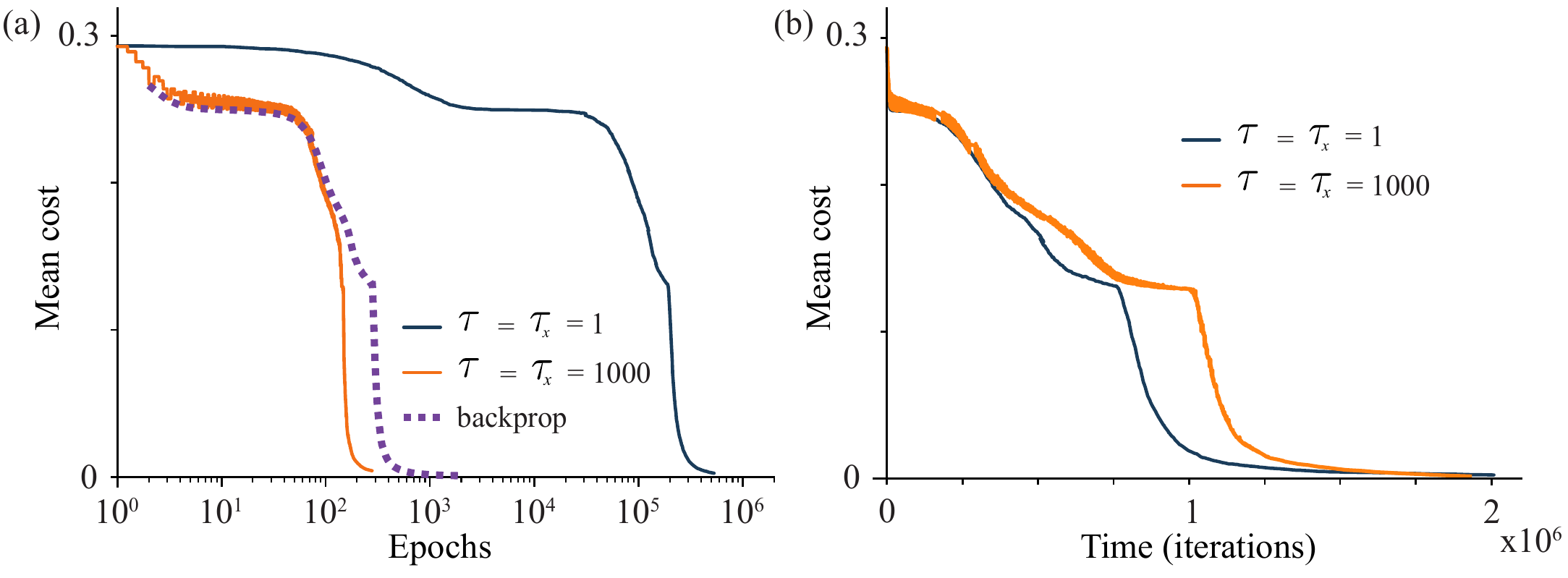}
    \caption{Solving the 2-bit parity (XOR) problem using a 2-2-1 network with 9 parameters, $\tau_p = 1$, and batchsize \tautheta/\taux = 1. (a) Mean cost over dataset versus epochs the on a 2-2-1 feedforward network averaged over 1000 random initializations and trained with MGD with different gradient integration times \tautheta (solid lines). The dashed line shows the same network and dataset trained using backpropagation. (b) The same MGD training data as in part (a) plotted versus time (number of perturbations \taup) instead of epochs. }
        \label{fig:4}
\end{figure}

\subsection{Equivalence to backpropagation}
\label{sec:simulation:equivalence}

We first verified that the simulation was able to minimize the cost for a sample problem, and that it was equivalent to gradient descent via backpropagation with appropriate parameter choices. For this initial comparison, we chose to solve the 2-bit parity problem by training a 2-2-1 feedforward network with 9 parameters (6 weights, 3 biases). We simulated the problem with a very large value for $\tautheta$ and $\tautheta = \taux$ such that we achieved a very good approximation of the gradient in $G$ for each training sample.  We then ran the same problem using a $\tautheta$ value of 1 so that the gradient approximation $G$ for each sample was relatively poor. We measured both the number of epochs and the amount of time (number of iterations of the simulation) for the two experiments, and the results are shown in \reffig{fig:4}. Here, an epoch was defined in the typical way – equivalent to the network being shown all training examples of a dataset.

Comparing the two scenarios in terms of epochs in \reffig{fig:4}a, one can observe that a value of $\tautheta = \taux = 1000$ resulted in the system following a nearly identical training trajectory as backpropagation. This is as expected -- for each sample shown to the network, the gradient approximation $G$ has 1000 timesteps to integrate a very accurate estimate that should be very close to the true gradient computed by backpropagation. When $\tautheta = \taux = 1$, however, each sample only has a single timestep to estimate the gradient before moving on to the next sample. As a result, the samples have to be shown to the network many more times to minimize the cost, resulting in a much larger number of epochs.

However, while the $\tautheta = \taux = 1$ case uses the sample data less efficiently (requiring more epochs),  there is actually a tradeoff for data efficiency and run time. If we plot the cost versus iterations rather than versus epoch, we get a more accurate estimate of how long it will take hardware to train in terms of real time.  As shown in \reffig{fig:4}b, one can see that the shorter \tautheta and \taux values take approximately half the time to minimize the cost as the longer values. These examples serve to highlight that while longer integration times produce a more accurate gradient approximation, integration times as short as \taup may also be used to train a network, consistent with the findings of \refcite{Spall1992}. In fact, shorter integration times may even be faster to train in terms of operational time.

\begin{figure}[h] % [H] forces the figure to appear exactly here
    \centering
    \includegraphics[width=3.5in]{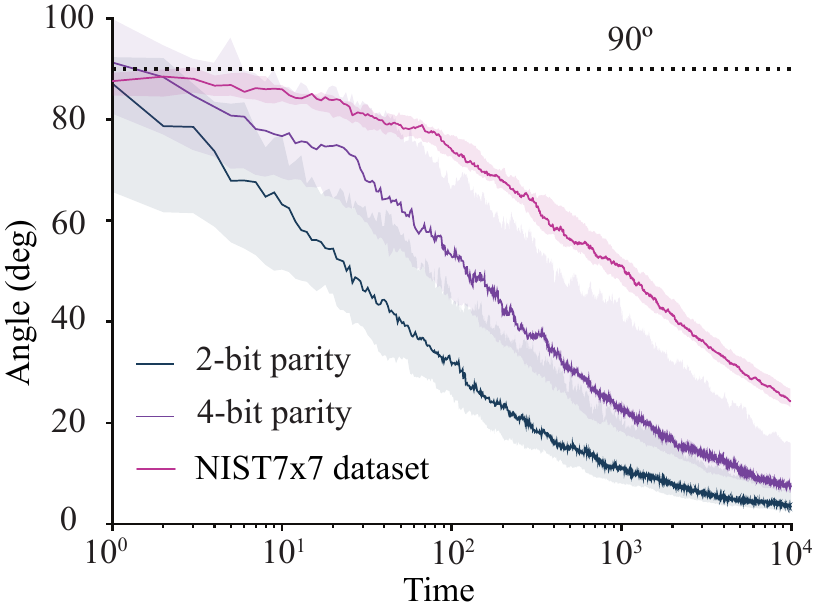}
    \caption{Angle between the gradient approximation $G$ and the true gradient versus time. The $n$-bit parity networks used $n$-$n$-1 networks, while the NIST7x7 networks used a 49-4-4 network . The networks have 9, 25 and 220 parameters respectively, and the datasets have 4, 16 and 44136 training examples respectively. The solid line shows the median angle value for 100 random initializations for the n-bit parity and and 15 random initializations for the NIST7x7 dataset. The shaded regions show the 1st to 3rd quartile values. }
        \label{fig:5}
\end{figure}

To quantify the effect of longer integration times on the accuracy of the gradient approximation, we measured how the gradient approximation $G$ converged to the true gradient $\partial C/\partial \theta$ (as computed by backpropagation) as a function of time. For this experiment, we ran the simulation with $\tautheta = \infty$ and $\taux = \taup = 1$, so that it continuously integrated $G$ without ever resetting or updating the parameters. As the simulation ran, we repeatedly computed the angle between the true gradient $\partial C/\partial \theta$ and the approximation $G$. The results are shown in \reffig{fig:5}, showing the solution to the 2-bit parity, 4-bit parity, and NIST7x7 problems. The NIST7x7 dataset is a small image recognition problem based on identifying the letters N, I, S, and T on a $7\times7$ pixel plane. The dataset has the property that it cannot be solved to greater than 93\% with a linear solve for a 49-4-4 feedforward network with sigmoidal activation functions. We also chose this network and dataset because it was small enough to perform many different simulations to acquire statistics, and the problem space is large enough that random solutions are unlikely.

As expected, the angle decreased with time as $G$ aligned with the true gradient. The time axis is in units of $\taup$, which is the minimum discrete timestep in this system. For a real hardware platform, this timestep is approximately the inference time of the system. In general, the more parameters the network has, the longer it takes to converge to the true gradient.

\begin{figure}[h] % [H] forces the figure to appear exactly here
    \centering
    \includegraphics[width=6.5in]{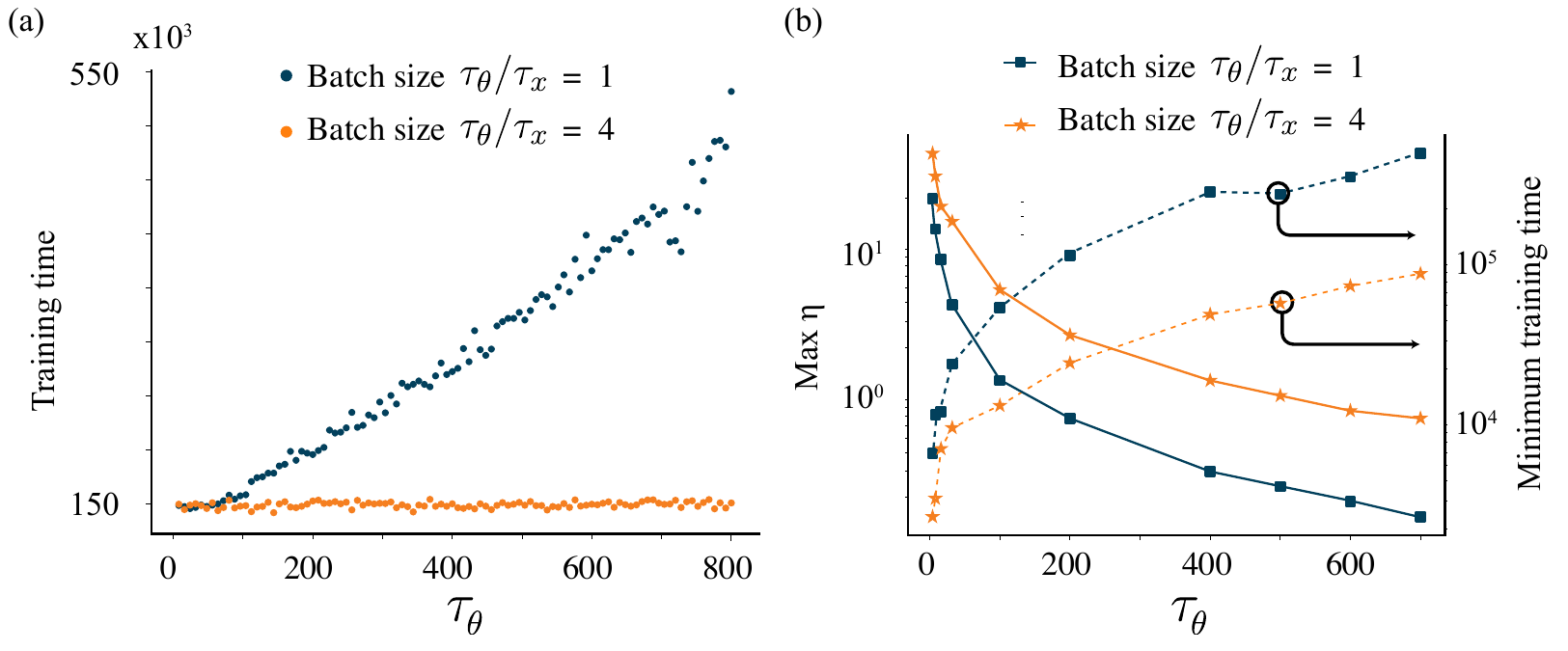}
    \caption{Effect of \tautheta on the training time of the 2-bit parity (XOR) problem. (a) Training time as a function of \tautheta and batch size.  (b) Effect of \tautheta on the maximum achievable learning rate $\eta$ and corresponding minimum training time.  }
        \label{fig:6}
\end{figure}

\subsection{Details of mini-batching}
\label{sec:simulation:details}

We next investigated in more detail how \tautheta and \taux affect the training time. Longer \tautheta values result in a more accurate gradient approximation, but reduce the frequency of parameter updates. Using a fixed, low $\eta$ value, we trained a 2-2-1 network to solve 2-bit parity (XOR) for 100 different random parameter initializations, varying \tautheta but keeping the batch size $\tautheta/\taux$ constant at either 4 or 1. Since the 2-bit parity dataset is composed of four ($x$, \yhat) pairs, $\taux = 4 \tautheta$ is analogous to gradient descent – all four samples are integrated into the gradient approximation $G$ before performing a weight update. When $\taux = \tautheta$, the network performs stochastic gradient descent (SGD) with a batch size of 1. \reffig{fig:6}a shows the training time as a function of \tautheta for these two cases. Here, training time corresponds to the time at which the total cost $C$ dropped below 0.04, indicating the problem was solved successfully. In the case where the batch size was 1, we observed that increasing \tautheta increased the training time. However, when the batch size was 4, increasing \tautheta had little effect on the training time. 

As with any training process, the training can become unstable at higher $\eta$ values and fail to solve the task. Since the results in \reffig{fig:6}a were only for a fixed learning rate, we also wanted to examine the effect of \tautheta on the maximum achievable $\eta$.  Here, we defined the ``max $\eta$'' as the maximum learning rate where the network successfully solved the 2-bit parity problem for at least 50 out of 100 random initializations. As seen in \reffig{fig:6}b, as \tautheta is increased the max $\eta$ decreases, resulting in longer minimum training times. This was true whether the batch size was large or small, although the larger batch sizes had higher achievable $\eta$ values overall.

From these results, we infer that a poor gradient approximation taken with respect to \textit{all} training examples is more useful than collecting an accurate gradient with respect to a single example. Stated another way, waiting a long time for an extremely accurate gradient then taking a large step is  less productive than taking a series of short (but less accurate) steps. This is an important conclusion for hardware, as it shows that implementing an effective gradient descent process in MGD does not necessarily require additional memory to store accurate, high-bit-depth gradient values. Note that in our implementation $G$ accumulates with time and so the size of the parameter update $\eta G$ from \refeq{eqn:update} grows proportionally to the integration time. This meant that when \tautheta was larger the effective step in the direction of the gradient was also larger, and so for fixed $\eta$ the rate of training therefore remains approximately constant. If this was not the case, whenever \tautheta was doubled we would also need to reduce $\eta$ by half to maintain the same approximate rate of training.

\subsection{Analog and digital perturbations}
\label{sec:simulation:analog}

The parameter perturbations can take many different forms, as long as they are low-amplitude and their time averages are pairwise orthogonal or, in a statistical setting, are uncorrelated~\cite{Dembo1990}. We have implemented four types of perturbations: sinusoidal perturbations, sequential discrete perturbations, discrete code perturbations, and random code perturbations.  Sinusoidal perturbations are like those shown in \reffig{fig:1-mgd}a, where each parameter is assigned a unique frequency.  Sequential discrete perturbations refer to the case where where parameters are sequentially perturbed, one at a time, by $+\Delta\theta$, as described in the finite-difference implementation of \refsec{sec:gradient}. ``Code'' perturbations are simultaneous discrete perturbations of $\{-\Delta\theta,+\Delta\theta\}$ for every parameter every \taup timesteps. We call them code perturbations due to their similarity to code-multiplexing (spread-spectrum) techniques used in wireless communication technologies. There are two flavors of code-perturbations: the first consists of a predefined set of pairwise-orthogonal square wave functions that take the values of $\{-\Delta\theta, +\Delta\theta\}$. Each of these perturbation patterns is a deterministic sequence, and no two parameters have the same sequence. One example of these are the Walsh codes used in modern cell-phone communications. The second consists of randomly-generated sequences of $\{-\Delta\theta, +\Delta\theta\}$ that are pairwise uncorrelated.  We call these ``statistically orthogonal.'' The statistically orthogonal case is slightly less efficient than the deterministic orthogonal codes since perturbations from multiple parameters interfere with each other more in \tc -- any finite sample of the perturbations will have a non-zero correlation that decreases to zero with sample size. However, the use of the statistically orthogonal version allows the perturbations to be generated locally and randomly. These perturbations may be very useful in hardware implementations, as they are  spread-spectrum and single-frequency noise from external sources is unlikely to corrupt the feedback signals. 

To compare the training performance between different perturbation types, we applied four different perturbation types to the same 2-2-1 network described in the last section in order to solve the 2-bit parity problem. In particular, we aimed to show that training can happen in both a purely analog or purely digital way. In practice, many systems have hybrid analog-digital features, and a combination approach may be used.

\begin{figure}[h] % [H] forces the figure to appear exactly here
    \centering
    \includegraphics[width=3.5in]{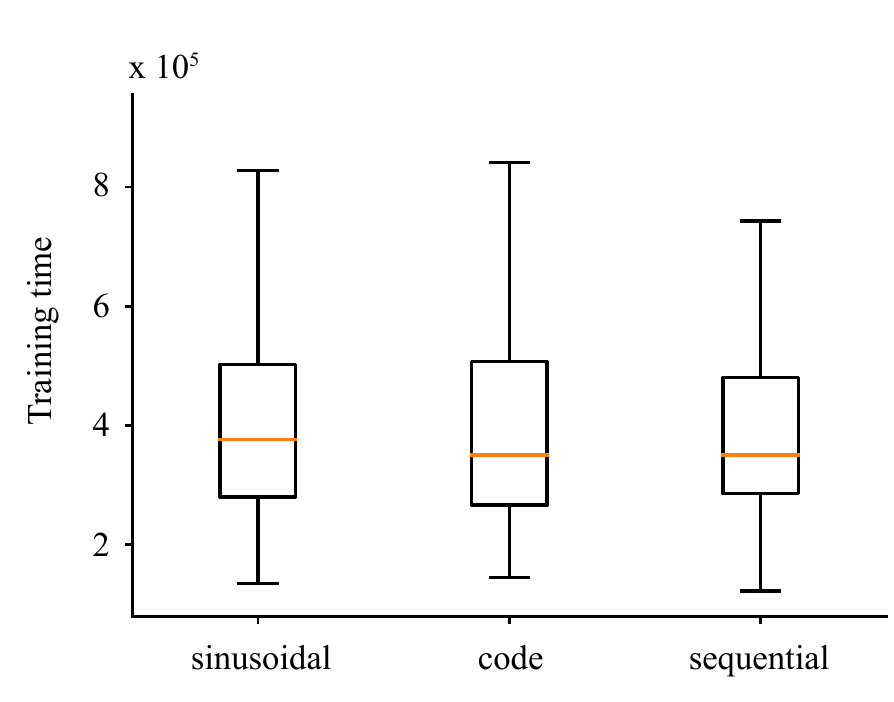}
    \caption{Demonstrating equivalence between the various perturbation types by measuring their time to train a 2-2-1 network on the 2-bit parity (XOR) problem. Each box plot represents the results of 100 random initializations. The hyperparameters were $\taux = 250$, $\tautheta = 1$, $\eta = 0.05$, $\taup = 1$ (discrete), $\Delta f = 0.3$ (analog)}. 
        \label{fig:7}
\end{figure}

\reffig{fig:7} shows the training time distributions for the 2-bit parity problem using the different perturbation types. The bandwidth for sinusoidal perturbations was set to be $1/2\taup$.  As expected, we found that the different perturbation types are approximately equivalent in terms of speed of training. This equivalence makes sense when one considers that the feedback from \tc has a finite bandwidth that must be shared between all the parameters -- no matter the encoding (perturbation) scheme, the information carried in that feedback to the parameters will be limited by that finite bandwidth.

\subsection{Operation on noisy or imperfect hardware}
\label{sec:simulation:operation}

The fabrication defects and signal noise present in emerging hardware platforms pose significant challenges for current training techniques in hardware. For example, weight parameters may differ from design or update non-deterministically, causing a discrepancy between the model and actual hardware that can negatively affect training \cite{Hoskins2021,Zhou2021,Wright2022}.  For example, in \refcite{Zhou2021}, a simulated network achieves 97.6\% accuracy for the MNIST dataset, but the hardware diffractive optical network implementation only obtains 63.9\% accuracy after direct transfer of the pre-trained model to their hardware without some in-situ training, while \refcite{Bandyopadhyay2021} shows that a 3\% component variation can lead to a 7\% drop in classification accuracy without error correction. \refcite{Wright2022} shows how small discrepancies in modeled parameters can blow-up in general hierarchical problems. In their toy problem, a 0.5\% error in a model parameter leads to a 30\% error in output after 20 layers. The solutions to these non-idealities can be cumbersome for offline training. For example, the actual value of the weights may have to be regularly measured during training \cite{Hoskins2021}, imperfections may be carefully measured and accounted for in the models \cite{Khosrowshahi2019,Bandyopadhyay2021}, or weights may be quantized with low bit-depth, where the bit depth is chosen such that the system can be modeled and controlled accurately despite device-to-device variations. This is common in hardware platforms such as memristive crossbar arrays \cite{Hu2018} or phase change materials \cite{Ambrogio2018}, where bit depths of 6-8 can be achieved.

Here we investigate the effects of three different types of imperfections and noise that could affect hardware systems: (1) stochastic noise on the output cost \Cnoise, (2) stochastic noise on the parameter update \thetanoise, (3) per-neuron defects in the activation function, where each neuron has a randomly scaled and offset sigmoidal activation function that is static in time.  These tests were performed on the NIST7x7 dataset using the previously described 49-4-4 network with 220 parameters and with $\taux = \tautheta = 1$ unless otherwise stated.

\begin{figure}[h] % [H] forces the figure to appear exactly here

    \centering
    \includegraphics[width=6.5in]{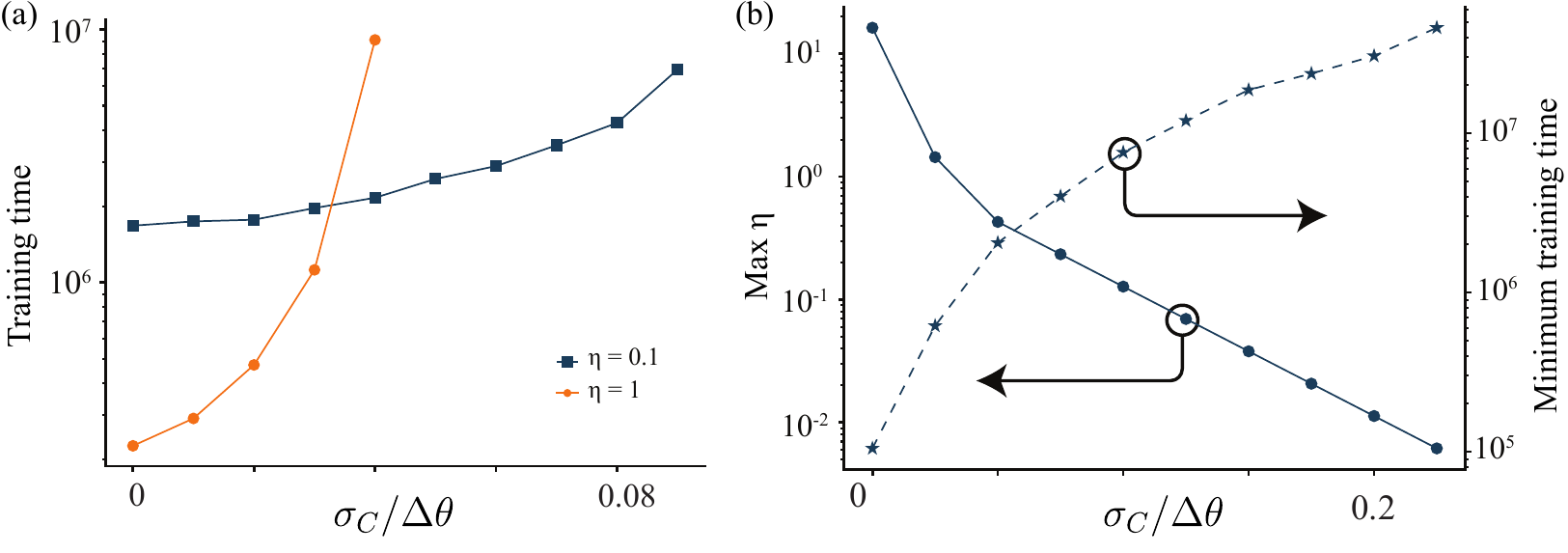}
    \caption{Effect of noise applied to the cost signal. (a) Training time versus \sigmac, the amplitude of the Gaussian noise added to the cost signal $C$ (normalized to the perturbation magnitude $|\ttheta|$). The training time is the median time to $>80\%$ accuracy for 10 random parameter initializations.  (b) Effect of noise on maximum $\eta$ and training time. The maximum $\eta$ is the largest learning rate for which $>80\%$ of the 10 random initializations converged; the corresponding training time at that maximum learning rate is shown on the right axis.}
        \label{fig:8}
\end{figure}

In the first test, we added Gaussian noise with mean zero and standard-deviation \sigmac to the cost,  applied every timestep such that noise $C(t) = C_{\text{ideal}}(t) + \Cnoise(t; \sigmac)$. For example, in optical hardware, this could be noise due to laser fluctuations. Noise in the cost is also equivalent to any other gaussian mean-zero noise affecting the accumulated gradient $G$.  \reffig{fig:8}a shows the effect on the training time for different learning rates as \sigmac was increased.   For a given learning rate, there is a threshold amount of noise below which the training time is minimally changed. However, as cost noise increases, the training time eventually increases and ultimately stops converging. 

To see how this noise would affect the minimum training time for optimized learning rates, we also measured the maximum achievable $\eta$ value for a range of \sigmac. \reffig{fig:8}b shows this maximum $\eta$ value versus cost noise, and corresponding minimum training time. The trend indicates that the lower the cost noise \sigmac, the higher we can make $\eta$ and the faster we can train.  Stated in reverse, one can compensate for large amounts of cost noise in a system simply by reducing the learning rate.

\begin{figure}[h] % [H] forces the figure to appear exactly here
    \centering
    \includegraphics[width=5in]{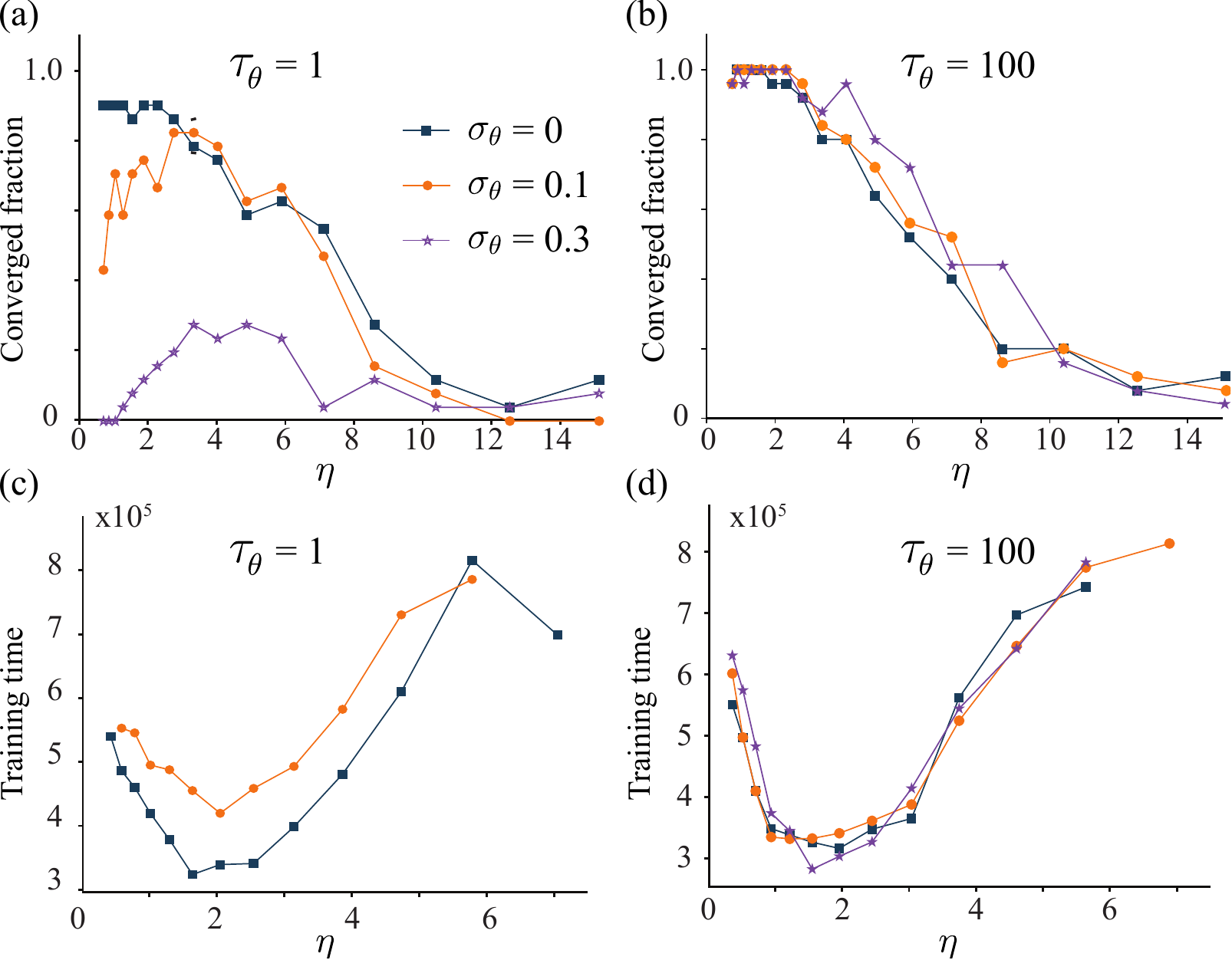}
    \caption{Effect of noisy (stochastic) parameter updates on solving XOR in a 2-2-1 feedforward network, measured for various noise amplitudes \sigmatheta. (a,b) Effect of parameter update noise on the probability of the training converging, as a function of learning rate $\eta$. Noise was much more likely to prevent convergence for (a) $\tautheta=1$ than (b) $\tautheta=100$.  Convergence was defined as reaching 93\% accuracy within  $5\times 10^7$ iterations.  Statistics are from 25 random parameter initializations. (c,d) Effect of parameter update noise on training time as a function of the learning rate $\eta$ for integration times (c) \tautheta = 1 and (d) \tautheta = 100. Only $\eta$ values sets with $>50\%$ convergence are shown -- in (c) \sigmatheta = 0.3 did not have $\eta$ values that met this criteria.}
        \label{fig:9}
\end{figure}

In the next test, we analyzed the effect of noisy parameter updates on the training process. For this experiment, whenever any parameter was updated, the update included a randomly-applied deviation. Thus, the update rule became

\begin{equation}
\label{eqn:noisy-update}
\begin{split}
\theta & \leftarrow \theta - \eta G + \thetanoise
\end{split}
\end{equation}

where \thetanoise was Gaussian with mean zero and standard deviation \sigmatheta, normalized by $\Delta \theta$, such that $\thetanoise \sim \mathcal{N}(0,\sigmatheta/\Delta \theta)$. This kind of noise is often seen in analog memories without closed-loop feedback \cite{Ambrogio2014,Hazra2019}.  

We found that larger values of \sigmatheta can prevent convergence entirely (\reffig{fig:9}a). Curiously, in the presence of this noise, increasing $\eta$ can actually improve the convergence of the problem, as highlighted by the $\sigmatheta = 0.1$ and $\sigmatheta = 0.3$ lines in \reffig{fig:9}a.  We believe this is likely because at very small $\eta$ values \thetanoise will overwhelm the the very small $\eta G$ in \refeq{eqn:noisy-update}.  By making the $\eta$ term larger, one can prevent $\eta G$ from being drowned out by \thetanoise.  Obviously, at very large $\eta$ values, the usual gradient-descent instability starts to dominate and the convergence approaches zero. For a given $\eta$, we found that small values of \sigmatheta marginally increase the training time, but the effect is less significant than simply changing the learning rate $\eta$ (\reffig{fig:9}c).

Another way to reduce the impact of \thetanoise is to increase the integration time of the gradient. When \tautheta is increased, the value of $G$ is accumulated for a longer time and becomes proportionally larger. For instance, when \tautheta is 100 times larger, the value of $\eta G$ becomes 100 times larger, meaning the effect of \thetanoise in \refeq{eqn:noisy-update} is relatively 100 times smaller. The result can be seen in \reffig{fig:9}b,d, where even the largest \sigmatheta value has little effect on the result. 

\begin{figure}[h] % [H] forces the figure to appear exactly here
    \centering
    \includegraphics[width=6.5in]{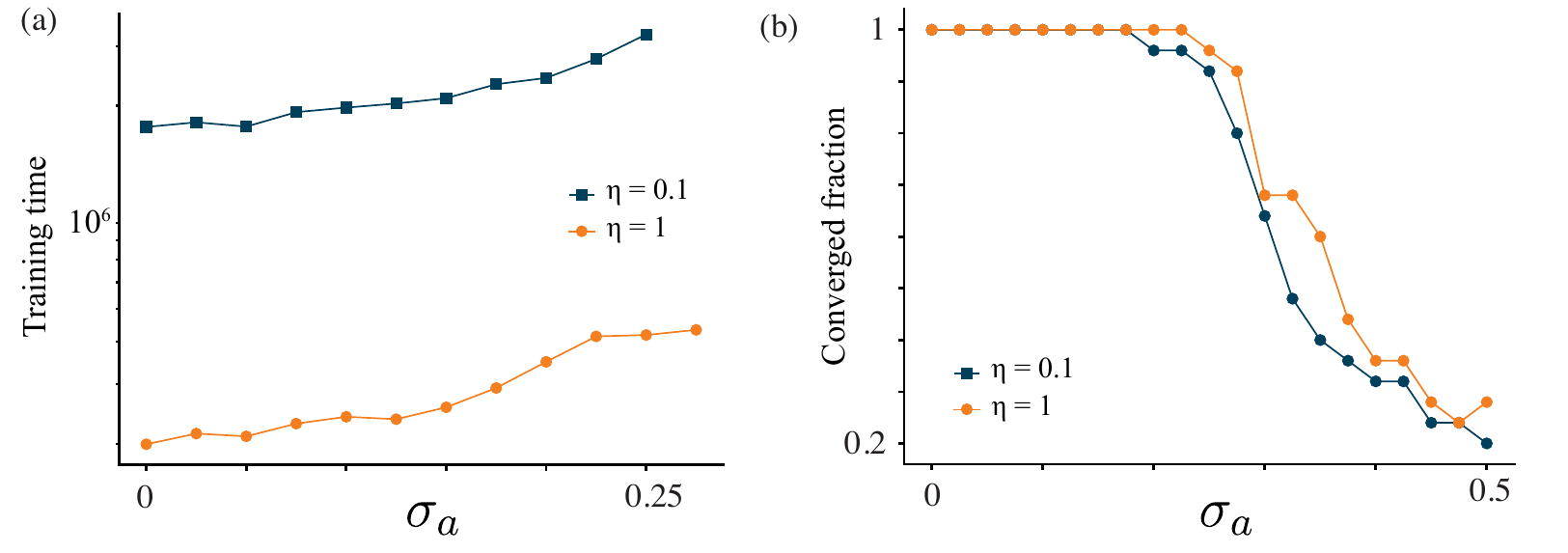}
    \caption{Effect of adding random offsets and scaling to each neuron's sigmoid activation function. (a) Training time versus $\sigma_a$, the standard deviation of activation-function offsets and scaling. (b) Converged fraction versus the standard deviation of the logistic function parameters. The results are statistics of 25 different random network initializations and activation-function randomizations. Convergence is defined as reaching 80\% accuracy within the time of the simulation. 
}
    \label{fig:10}
\end{figure}

In the final test, we analyzed the effect of including ``defects'' in the neuronal activation functions. Here, neuronal activation functions were no longer identical sigmoid functions, but had fixed random offsets and scaling that were static in time. These variations were meant to emulate device-to-device variations that may be found in hardware, for instance in analog VLSI neurons \cite{Merkel2015}. The sigmoid activation function for each neuron $k$ was modified to a general logistic function $f_k(a) = \alpha_k(1-e^{-\beta_k(a-a_k)})^{-1} + b_k$. The variations were all Gaussian, and the scaling factors $\alpha_k$, $\beta_k$ had a standard deviation $\sigma_a$ and a mean of 1, while the offset factors $a_k$, $b_k$ also had a standard deviation of $\sigma_a$ but were mean-zero.

As seen in \reffig{fig:10}a, adding defects to the network's activation functions had a relatively small effect on the training time.  Even with relatively large variations in the activation functions ($\sigma_a = 0.25$), the network only took about twice as much time to fully train the NIST7x7 dataset.  However, deviations that were too large ($\sigma_a > 0.25$) caused issues with the training being able to converge.  This is likely due to the output neurons being so scaled and offset that it was no longer possible for them to produce the correct output.

Overall, we found that MGD is robust against different types of noise and fabrication imperfections of a reasonable scale as those that would be seen in a real hardware system.  In general, these non-idealities affected the training speed of the network, but did not prevent the desired problem from being solved.

\subsection{Modern dataset results}
\label{sec:simulation:modern}

\begin{table}[]

\begin{adjustbox}{width = 1\textwidth}
\small
\begin{tabular}{l|l|l||r|r|l|r||r|r|r|r|r}
\multicolumn{2}{c}{\textbf{Setup}} & \multicolumn{5}{c}{\textbf{Parameters}} & \multicolumn{4}{c}{\textbf{Accuracy}} \\ 
\textbf{Task} & \textbf{Network} & \textbf{$|\theta|$} & \textbf{\tautheta} & \textbf{\taup} & \textbf{$\eta$} & \textbf{batch size} & \textbf{$10^{4}$ steps} & \textbf{$10^{5}$ steps} & \textbf{$10^{6}$ steps} & \textbf{$10^{7}$ steps} &\textbf{backprop} \\ \hline
2-bit parity & 2-2-1 & 9 & 1 & 1 & 5 & 1 & 100\% & 100\% & 100\% & 100\% & 100\% \\
N-I-S-T & 49-4-4 & 220 & 1 & 1 & 3 & 1 & 38\% & 81\% & 94\% & 97.7 \% & 99.8\% \\
N-I-S-T & 49-4-4 & 220 & 1 & 1 & 0.5 & 1 & 22\% & 45\% & 93\% & 98.7 \% & 99.8\% \\
Fashion-MNIST & 2-layer CNN & 14378 & 1 & 1 & 9 & 1000 & 34.2\% & 66.3\% & 79.3 \% & 83.5\% & 88.6\% \\
Fashion-MNIST & 2-layer CNN & 14378 & 10 & 1 & 9 & 1000 & 34.3\% & 66.3\% & 79.2 \% & 83.4\% & 88.6\% \\
Fashion-MNIST & 2-layer CNN & 14378 & 100 & 1 & 9 & 1000 & 35.3\% & 66.3\% & 77.7 \% & 84.7\% & 88.6\% \\
Fashion-MNIST & 2-layer CNN & 14378 & 1000 & 1 & 9 & 1000 & 35.3\% & 59.6\% & 79.1 \% & 86.1\% & 88.6\% \\
CIFAR-10 & 3-layer CNN & 26154 & 1 & 1 & 9 & 1000 & 12\% & 23\% & 43.8\% & 60.7\% & 68\% \\
\end{tabular}
\end{adjustbox}
  \caption{Performance of MGD training on four different datasets. The setup, MGD parameters and achieved test accuracies are shown. The best accuracy achieved with training via backpropagation for the same network is shown in the final column for comparison.}
  \label{table:results}
\end{table}

To assess the scalability and relevance of MGD for larger machine learning problems, we compared MGD and backpropagation on a variety of tasks for different network architectures and hyperparameters. \reftab{table:results} compares the accuracies obtained with MGD and backpropagation for different datasets and various hyperparameter choices (\tautheta, \taup, $\eta$, batch size), with \taux fixed at 1.  To make an effective comparison, for the backpropagation results we used a basic stochastic gradient descent (SGD) optimizer without momentum. Both strategies used a mean squared error (MSE) cost function. These choices allowed us to avoid confounding effects from more complex strategies, although we note that MGD is capable of implementing several of these more complicated feature (e.g., momentum, dropout, etc.). In all cases we used the same network architectures for both sets of results. The networks architectures used were not selected to reach state-of-the-art accuracy values, but instead smaller networks were chosen for purposes of rapid testing and statistical analysis. To improve the accuracy further, more advanced network architectures (e.g. more layers) or optimizers (e.g. adding momentum) would be needed. For the purposes of meaningful comparison, we measured the accuracy obtained in the MGD process after fixed numbers of timesteps. This allowed us to estimate actual training time for various hardware configurations in \refsec{sec:hardware:time}.

The CIFAR-10 network was composed of 3 convolution and max-pool layers followed by a fully-connected layer.  The convolutional filters were $3 \times 3$ with 16, 32 and 64 output channels respectively, stride 1, and relu activation function outputs.  Each convolution was followed by a $2\times 2$ max-pool layer. The final fully-connected layer converted the 256 outputs of the convolutional layers to the 10 classes, and no softmax was used. The Fashion-MNIST network was similarly composed, comprising two convolution and max-pool layers followed by a ($32 \times 10$) fully-connected layer and no softmax. The networks used to solve NIST7x7 and XOR were fully connected feedforward networks with sigmoidal activation functions. To make the comparisons in \reftab{table:results} as fair as possible, the backpropagation accuracies of each row were maximized by training the networks with the listed batch size and a variety of $\eta$ values.  The backpropagation accuracy reported is the highest median accuracy obtained for all the $\eta$ values tested, with the median taken over five random initializations that each ran for 2500 epochs (long after the accuracy had converged).

By training networks on the 2-bit parity, NIST7x7, Fashion-MNIST, and CIFAR-10 datasets, we found that MGD approached the solution-accuracy of backpropagation.  However, even after $10^7$ steps we observed the MGD-trained networks still had slightly lower accuracy than could be achieved by backpropagation.  This is likely related to convergence criteria of the gradient approximation not being met, due to fixed $\eta$ values – for instance, the theory underlying the SPSA process only guarantees convergence with a learning rate that asymptotically approaches zero \cite{Spall1992}. In general, we observed that while larger learning rates achieved higher accuracies earlier on, lower learning rates achieved higher accuracies at longer times (see e.g. NIST7x7 result).  Therefore custom learning rates are likely to achieve more optimal training time and accuracy.

The simulations also indicate that changing \tautheta had a marginal effect on the maximum accuracy of MGD for a fixed $\eta$ value for the larger datasets shown here, with increased values of \tautheta leading to slightly higher final accuracies. This may be because longer \tautheta values produce more-accurate gradient estimates and allow the network to better optimize the weights as the system approaches a local minima in the cost, although more experiments and statistics are needed to verify if this effect is occurs in general.  Fortunately, our results show this effect is relatively small, allowing the hardware-designer flexibility in how often weight updates need to be performed. 

These results show explicitly that MGD is a viable technique for training emerging hardware platforms on real machine learning datasets. Previous theoretical studies have focused on convergence proofs and computation of gradient approximations to varying degrees of accuracy \cite{Spall1992,Dembo1990} or on simplified problems, such as linear activation functions \cite{Werfel2005}. While these theoretical results are clearly important and necessary, they are not sufficient to give insight into the applicability of these optimization techniques to real machine learning problems on experimental hardware platforms. Meanwhile, previous experimental and simulation studies have for the most part focused on small scale problems, for which 100\% accuracy is trivial, such as XOR \cite{Matsumoto1990,Jabri1992}, few-bit parity \cite{Alspector1992,Miyao1997,Montalvo1997},  the iris dataset \cite{Wang2018} or other similar toy problems \cite{Kirk1992,Maeda1995,Bandyopadhyay2022}. These are not representative of larger machine learning datasets, which typically have on the order of tens of thousands of training examples and are trained on networks with even more parameters.

These results go further than any of these previous studies, by elucidating the utility of these techniques for larger, modern, machine learning datasets, and implementing a framework that can connect different optimization techniques (SPSA, finite-difference, approximate gradient descent) via specific hardware parameters.

It is also worth noting that the random weight change (RWC) algorithm  has been implemented in memristor networks \cite{Adhikari2015, Yang2016} previously, and while superficially similar to MGD is functionally very different. RWC is not an approximate gradient descent technique, since the weight update is not scaled by the magnitude of the change in the cost, but rather random perturbations are either kept or discarded based on whether or not they improve the cost. Because of this, it scales more poorly with number of parameters than the optimization techniques we have implemented in the MGD framework. 

\section{Hardware considerations}
\label{sec:hardware}

\subsection{Time constants}
\label{sec:hardware:time}

\begin{table}[ht!]
\begin{adjustbox}{width = 1\textwidth}
\small
\begin{tabular}{l|l|l|l|l}
 & \textbf{HW1} & \textbf{HW2} & \textbf{HW3} & \textbf{backprop} \\ \hline
\textbf{\taux (input-sample update time)} & 100 ns & 1 ns & 10 ps & n/a \\
\textbf{\taup (perturbation time)} & 1 ms & 10 ns & 200 ps & n/a \\
\textbf{\tautheta (parameter-update time)} & 1 ms & 1 \textmu s & 200 ps & n/a \\
% \textbf{Batch size} & 1000 & 1 &  \\
\textbf{2-bit parity training time ($10^4$ steps)} & 20 s & 200 \textmu s & 4 \textmu s &  70 ms$^\dagger$ \\
% \textbf{NIST training time} &  & &  \\
\textbf{Fashion-MNIST training time ($10^6$ steps)} & 33 min  & 20 ms & 400 \textmu s &  54 s$^*$ \\
\textbf{CIFAR-10 training time ($10^7$ steps)} & 5.6 hours & 200 ms & 4 ms &  480 s$^*$\\ \hline
\textbf{Examples of hardware} &\makecell[l]{chip-in-the-loop, \\integrated photonics \\with thermo-optic tuning\cite{Tait2017,Shen2017}} & \makecell[l]{Mem-compute devices \cite{Burr2017},\\ analog VLSI \cite{Kohda2020}} & \makecell[l]{superconducting devices \cite{Schneider2022},\\ athermal resonant \\silicon photonic modulator \cite{Timurdogan2014}} & \makecell[l]{$^\dagger$CPU/$^*$GPU} \\
\end{tabular}
\end{adjustbox}
\caption{Approximate training time using MGD for estimated hardware time constants. Data shown is estimated using the same dataset and network architectures as used in \reftab{table:results}. The final column shows the time it takes to get to that same accuracy using backpropagation on a standard GPU or CPU.}
\label{table:hardware}
\end{table}

While the MGD framework is general, individual hardware platforms may have different practical considerations that dictate the optimal choices for the different time constants. Here we discuss potential issues for implementation on hardware, and point out tradeoffs that may exist.

\textbf{Perturbation speed (\taup):} In many hardware platforms, perturbing parameters directly may be difficult or undesired either due to memory-endurance or update-speed concerns. However, it is not necessary to to perturb parameters directly--instead, one may perturb a separate element that is in series with the parameter. For instance, in a photonic MZI array, the weights may be implemented with slow ($\sim$millisecond) thermal or mechanical phase shifters, but the small perturbation could be implemented with a fast ($\sim$nanosecond) electro-optic modulator such as a depletion \cite{Timurdogan2014} or opto-mechanical \cite{Dong2022} modulator. Alternatively, in a memristor system, the perturbation could be implemented with a single sub-threshold transistor in series with the memristor.

For perturbative-style techniques like MGD to function properly, \taup should be slower than the system's inference time \cite{Dembo1990} which includes the cost calculation and broadcast processes. If \taup is too short, it can introduce misalignment between the input perturbation and resulting cost feedback.

\textbf{Parameter update (\tautheta);} When performing gradient descent on hardware, parameters must be updated multiple times. For a training period of length $T$, the weights will be updated $T/\tautheta$ times. For some hardware platforms it may be advantageous to reduce the number of parameter updates, for instance if parameters have a limited lifetime, take a long time to update, or the update process requires a large amount of energy.
Fortunately, \reftab{table:results} shows that for networks with large numbers of parameters, \tautheta can be increased without greatly impacting the overall training speed or final accuracy.

\textbf{Changing training examples (\taux):} The value of \taux is limited by the speed with which training examples can be input to the network. For most technologies, \tautheta or \taux will not be limiting factors, as $x$ and \yhat samples can be provided by a conventional computer or FPGA at nanosecond timescales. However, it should be noted that MGD will not function correctly if \taux is shorter than the inference time of the hardware, and this will likely be the practical limit for \taux. \taux is also used as a way of controlling the batch size (given by $\tautheta/\taux$). The particulars of the task and dataset may determine the desired batch size, and this may have some effect on \taux, for example if the minimum allowable value of \tautheta is very long.

\textbf{Calculation of cost:} The calculation of the cost is performed only at the output of the system, and therefore it is acceptable that the hardware used in its computation be more expensive and complex. In integrated electronic systems, the cost can be straightforward to implement, although it may occupy more relatively more chip real-estate than other elements.  However, for very exotic hardware platforms, this cost computation may still be difficult to perform on-chip. This can be addressed by using a non-standard cost function or by using a computer to calculate the cost off-chip. However, the input and output to an off-chip computation may limit the speed of the cost computation and also limit the speed of the global broadcast of \tc, and may be another speed bottleneck in some hardware systems. 

Based on the literature, some estimates of plausible time constants that could be implemented in hardware are shown in \reftab{table:hardware} with the corresponding times to solve benchmark tasks, based on the results from \reftab{table:results}.Examples from the literature of hardware platforms that could implement these sets of parameters are shown in the final row. By comparison of these results with the final column in \reftab{table:hardware}, we see that using realistic estimates for emerging hardware, MGD could be significantly faster than current implementations using backpropagation.

\subsection{Analog and digital implementations}
\label{sec:hardware:analog}
There are a few notable differences for an analog versus a digital hardware implementation of the MGD framework. The discrete case requires one memory element at the network output to store $C_0$ (sample-and-hold), and a simple subtraction operation to compute $\tc = C-C_0$.  The network also requires some timesteps to be devoted to the measurement of $C_0$.  This means that for the simple case of $\tautheta = \taup$, only a single additional memory element is required for training the entire hardware system, located at the network cost output. 

However, in the case where $\tautheta > \tau_p$, an additional memory element is required for every parameter (for instance, an analog capacitor or discrete memory) to perform the gradient integration. The analog case requires a lowpass filter at every parameter element, and a single highpass filter on the network output to convert $C$ to $\tc$. These could be created with simple analog circuits like RC or LR filters. However, we note the use of a continuous highpass filter can cause implementation issues is some cases. For instance, if $\eta$ is too large, rapid changes in $\theta$ can generate unwanted frequency components that mix with the perturbation input and may negatively affect the gradient approximation. Moreover, if the dataset is discrete, jumps in $x$ can propagate high frequency noise through $C$ and $\tc$, again negatively affecting the gradient approximation. The simulations demonstrate that in principle MGD can be used on different hardware platforms of either an analog or digital nature for network training.

\section{Discussion}
\label{sec:discussion}
An interesting analogy can be drawn between the MGD training framework and wireless communication systems. In wireless communications, cell phone users must be able to extract the signal that is relevant to their own communication from a received broadcast that contains multiple users' signals. MGD operates similarly--each parameter is analogous to an individual cell phone user and the global cost broadcast \tc is analogous to the cell tower transmitter. The broadcasted signal \tc is available to all parameters, and each parameter extracts its relevant (gradient) information – analogous to voice data – from that broadcasted signal.

The encoding and multiplexing techniques used in wireless communications are broadly termed "multiple access" techniques in communications, and there are several varieties of them including frequency, time, and code multiplexing.  Similarly, in MGD, these multiplexing techniques are directly analogous to the different perturbation types discussed in \refsec{sec:simulation:analog}.  For instance, sinusoidal perturbations can be considered a type of frequency-multiplexing: each ``user'' (parameter) has a unique frequency containing information relevant to it, and must perform a time-integration to extract that information from the \tc signal which contains many other, irrelevant signals. Likewise, as mentioned in \refsec{sec:simulation:analog}, the discrete $\{-\Delta\theta,+\Delta\theta\}$ perturbation type is analogous to the code-multiplexing techniques used in modern cell phones.

It is important to note that for a fixed bandwidth broadcast, all multiplexing techniques have the same nominal spectral efficiency. However, the particulars of real systems can greatly affect which multiplexing techniques are most effective. For example, because code-multiplexing encodes a user's signal out over a wide time window and a spread spectrum, it has been shown to have an improved robustness to multipath distortion and timing errors compared to time-multiplexing.

Similarly, the various perturbation types described earlier may operate better or worse depending on the particulars of the hardware network and environment.   Newly developed communications techniques may also be directly applicable to MGD training in real hardware systems. Key to its usage in hardware, each parameter can operate like a cell phone –- as an independent device with a receiver, processor and memory using only information available locally and from the global broadcast. This contrasts with most other training algorithms, where error signals must propagate through the entire network \cite{Lillicrap2020}. While trivial in software, this type of upstream information-flow means that hardware neurons must (1) have additional memory to process the error signal, (2) perform different operations for forward-inference and reverse error propagation, and (3) operate in a synchronized fashion to propagate the error correctly.

In systolic arrays “broadcast” is often discouraged because the time and energy of communication is proportional to the number of inputs by CV$^2$ (with C growing with the number of inputs). Furthermore, as the capacitance is larger, then a larger amplifier is needed for the communication which also requires energy and area. However in other types of analog or emergent hardware, broadcast may be a more energy and space efficient process. For example, in optical hardware, a global optical signal could be accessible in a simple slab waveguide mode. Alternatively, in analog electrical hardware, a single electrical plane could be used to carry the \tc signal back to the individual parameters.

The MGD framework may also be much more biologically-plausible than other algorithms. There are a host of algorithms being developed to address the mystery of how the brain learns, and to use biology as inspiration to find more hardware-friendly approaches \cite{Lillicrap2020,Zenke2021,Scellier2017}. One of the major issues with backpropagation from both a practical and bio-plausibility standpoint is the need for a separate type of operation in the forward (inference) and backward (learning) phases. This is not the case for MGD: in MGD, the training is done in the same configuration used for inference. In a biological analogy, the a global cost signal \tc could be modeled as some type of quasi-global chemical signal in the brain.

In future, the MGD algorithm could be extended such that the cost is not global to every synapse, but rather to specific clusters of neurons/synapses. There are chemical signals in the brain with these types of properties. The time scales needed for perturbation may line up with biologically observed effects such as short-term plasticity (minutes) \cite{Soltoggio2015,Gerstner2018} and synaptic noise \cite{Rowland2007}. MGD is also consistent with ``three-factor'' learning rules, for which there is mounting experimental evidence \cite{Gerstner2018}.  Additionally, although not explored in this work, MGD should work in a spiking model. Similar algorithms that use perturbations in neuron conductance \cite{Fiete2006}, probabilistic synapse transmission \cite{Seung2003} or the stochastic nature of spike-timing arrival \cite{Xie2004} for reward-based learning have been explored in small spiking networks.

\section{Conclusions}
\label{sec:conclusions}
In this paper we show that with realistic timescales for emerging hardware, training via MGD could be orders of magnitude faster than backpropagation in terms of wall-clock time-to-solution on a standard GPU/CPU. The MGD framework allows the implementation of multiple optimization algorithms using a single, global, cost signal and local parameter updates.
The style of algorithm used (e.g. finite-difference, coordinate-descent, SPSA, etc.) can be adjusted via the tuning of the MGD time-constants, and can even be adjusted during training if desired. Because it is a model-free perturbative technique (sometimes called zeroth order optimization), it is applicable to a wide range of systems -- it can be applied to both analog and digital hardware platforms, and it can be used in the presence of noise and device imperfections. This overcomes a major barrier to using hardware platforms based on emerging technologies, which are often difficult to train.

Going forward, there are many opportunities to explore the application of MGD on both large and small hardware systems. The most direct next step will be to test the training performance in existing hardware, for example on photonic \cite{Tait2017} or memristive crossbar hardware using a chip-in-the-loop process.  This can occur without even redesigning the hardware, as MGD only requires the ability to input samples, capture inference output, and modify parameters when in a chip-in-the-loop configuration.  When used in this fashion, the speed will most likely be limited by system I/O. For example, perturbations can be injected directly to the hardware from an external computer, and that same computer could capture the changes in cost and perform the gradient approximation and calculate the weight updates.  This would allow testing of the algorithm without any changes to the hardware. Ultimately, to overcome I/O limitations local circuits can be designed be implemented for autonomous online training. Although not examined in detail here, in this case many of the hardware platforms examined would likely have several orders of magnitude improvement in terms of energy usage as well. 

There is also lots of room for improvement on tuning the technique and its hyper-parameters.  In this paper we performed very little hyperparameter optimization or regularization, and so there are likely opportunities for further optimization by examining techniques such as dropout, momentum, and regularization. Also of interest is examining the node-perturbation version of the algorithm \cite{Flower1992} on large datasets, which could significantly reduce the number of perturbative elements and speed up the training process. 

While in this paper we only examined feed forward networks, it has already been demonstrated that is possible to use perturbative techniques to train recurrent neural networks \cite{Cauwenberghs1992}, spiking networks \cite{Seung2003} and other non-standard networks at small scale; future work remains to demonstrate their utility on problems of modern interest. This will have applicability to a wide range of neuromorphic hardware and other physical neural networks.

Ultimately, MGD is well-suited to implementation directly on-chip with local, autonomous circuits. Although significant work remains before autonomous learning can be truly implemented without the participation of an external computer, such a device would be truly enabling for remote and edge-computing applications, allowing training in-the-field on real-time data.

%%%%%%%%%%%%%%%%%%%%%%%%
%%% Acknowledgements %%%
%%%%%%%%%%%%%%%%%%%%%%%%

\section{Acknowledgements}

The authors acknowledge helpful discussions with the NIST NCSG community.  This research was funded by NIST (https://ror.org/05xpvk416) and University Colorado Boulder (https://ror.org/02ttsq026).

%%%%%%%%%%%%%%%%%%%%%%%%%%%%%%%%%%%%
%%%%%%%%% Bibliography %%%%%%%%%%%%%
%%%%%%%%%%%%%%%%%%%%%%%%%%%%%%%%%%%%

\end{document}